\documentclass[a4paper]{ifacconf}

\usepackage{graphicx,amsmath,url}      
\usepackage[round]{natbib}   
\usepackage[table]{xcolor}
\usepackage{amsmath,amssymb}
\usepackage{xcolor}
\usepackage{mathtools}
\usepackage{xcolor}
\usepackage{collcell}
\usepackage[table]{xcolor}
\definecolor{lightgray}{gray}{0.96}
\usepackage{icomma}

\newenvironment{englishalgorithm}[1][]
  {\begin{algorithm}[#1]
     \selectlanguage{english}%
     \floatname{algorithm}{Algorithm}%
  }
  {\end{algorithm}}

\usepackage{multirow}

\usepackage[Algoritmo]{algorithm}
\usepackage[noend]{algpseudocode}
\usepackage{etoolbox}
\errorcontextlines\maxdimen

\makeatletter
\newcommand*{\algrule}[1][\algorithmicindent]{\makebox[#1][l]{\hspace*{.5em}\vrule height .75\baselineskip depth .25\baselineskip}}%

\newcount\ALG@printindent@tempcnta
\def\ALG@printindent{%
    \ifnum \theALG@nested>0
        \ifx\ALG@text\ALG@x@notext
            \addvspace{-3pt}
        \else
            \unskip
            \ALG@printindent@tempcnta=1
            \loop
                \algrule[\csname ALG@ind@\the\ALG@printindent@tempcnta\endcsname]%
                \advance \ALG@printindent@tempcnta 1
            \ifnum \ALG@printindent@tempcnta<\numexpr\theALG@nested+1\relax
            \repeat
        \fi
    \fi
    }%
\usepackage{etoolbox}
\patchcmd{\ALG@doentity}{\noindent\hskip\ALG@tlm}{\ALG@printindent}{}{\errmessage{failed to patch}}
\makeatother
\usepackage{tikz}
\usepackage{collcell}
 
\newcommand*{\MinNumber}{0.6449}%
\newcommand*{\MaxNumber}{1}%
 
\newcommand{\ApplyGradient}[1]{%
        \pgfmathsetmacro{\PercentColor}{100.0*(#1-\MinNumber)/(\MaxNumber-\MinNumber)}
        \hspace{-0.33em}\colorbox{gray!\PercentColor!black}{}
}
 
\newcolumntype{R}{>{\collectcell\ApplyGradient}c<{\endcollectcell}}
\renewcommand{\arraystretch}{0}
\def\portugues{1} 

\ifx\portugues\undefined
\def\portugues{0}
\fi

\if\portugues0
   \usepackage[english]{babel}
  \else
   \usepackage[spanish,brazil,english]{babel}
\fi

\usepackage[T1]{fontenc}
\usepackage[utf8]{inputenc}
\usepackage{ae}

\if\portugues1
 { }
 { }
 { }
 
\fi

\begin{document}

\if\portugues1

%
\selectlanguage{english}
	
\begin{frontmatter}

\title{Hybrid Method Based on NARX models and Machine Learning for Pattern Recognition} 


\author[First]{Pedro H. O. Silva} 
\author[First]{Augusto S. Cerqueira} 
\author[Second]{Erivelton G. Nepomuceno}

\address[First]{Department of Electrical Engineering, Federal University of \\ Juiz de Fora (UFJF), Juiz de Fora, MG, Brazil, \\ (e-mail: silva.pedro@engenharia.ufjf.br, 
augusto.santiago@ufjf.edu.br).}
\address[Second]{Department of Electrical Engineering, Federal University of São João del-Rei (UFSJ), São João del-Rei, MG, Brazil, \\ (e-mail: nepomuceno@ufsj.edu.br).}

\selectlanguage{english}
\renewcommand{\abstractname}{{\bf Abstract:~}}
\begin{abstract}                
This work presents a novel technique that integrates the methodologies of machine learning and system identification to solve multiclass problems. Such an approach allows to extract and select sets of representative features with reduced dimensionality, as well as predicts categorical outputs. The efficiency of the method was tested by running case studies investigated in machine learning, obtaining better absolute results when compared with  classical classification algorithms.

\vskip 1mm
\selectlanguage{brazil}
{\noindent \bf Resumo}: O presente trabalho apresenta uma nova técnica que integra as metodologias de aprendizado de máquinas e identificação de sistemas na solução de problemas multiclasses. A abordagem permite extrair e selecionar conjuntos de características representativas com dimensionalidade reduzida, da mesma forma que prediz saídas categóricas. A eficiência do método é testada pela  aplicação em estudos de casos estudados no aprendizado de máquina, obtendo melhores resultados absolutos em comparação aos algoritmos clássicos de classificação.

\end{abstract}


\selectlanguage{english}


\selectlanguage{english}

\end{frontmatter}
\else
%

\begin{frontmatter}

\title{Style for SBA Conferences \& Symposia: Use Title Case for
  Paper Title\thanksref{footnoteinfo}} 

\thanks[footnoteinfo]{Sponsor and financial support acknowledgment
goes here. Paper titles should be written in uppercase and lowercase
letters, not all uppercase.}

\author[First]{First A. Author} 
\author[Second]{Second B. Author, Jr.} 
\author[Third]{Third C. Author}

\address[First]{Faculdade de Engenharia Elétrica, Universidade do Triângulo, MG, (e-mail: autor1@faceg@univt.br).}
\address[Second]{Faculdade de Engenharia de Controle \& Automação, Universidade do Futuro, RJ (e-mail: autor2@feca.unifutu.rj)}
\address[Third]{Electrical Engineering Department, 
   Seoul National University, Seoul, Korea, (e-mail: author3@snu.ac.kr)}
   
\renewcommand{\abstractname}{{\bf Abstract:~}}   
   
\begin{abstract}                
These instructions give you guidelines for preparing papers for IFAC
technical meetings. Please use this document as a template to prepare
your manuscript. For submission guidelines, follow instructions on
paper submission system as well as the event website.
\end{abstract}


\end{frontmatter}
\fi


\section{Introduction}

The progressive development of modern technology, comprised of computer and internet applications, generates large amounts of data at an unprecedented speed, such as videos, photos, texts, voices, and data obtained from the emergence of the Internet of Things (IoT) and cloud computing. Data often have large attributes, presenting sets of redundant, noisy, and irrelevant features that can degrade the performance of machine learning algorithms, posing a major challenge for data analysis and decision making. Therefore, there is a need to use techniques that allow reducing the dimensionality of data, which is a step that helps data mining and machine learning algorithms to be more efficient.

The dimensionality reduction problem can be solved using feature extraction techniques. Feature extraction deals with the problem by generating a new reduced set of features with $k$ dimensions, coming from combinations of the original set with $d$ dimensions \citep{Cai2018}. The new reduced dataset has high discriminatory power, which can increase algorithm performance, reduce processing time, and simplify results. Furthermore, in some cases, feature extraction promotes an increase in the understanding of the results and leads to an improvement in precision, as it avoids excessive adjustments to the data sample.

In machine learning, conventional classifiers largely lack processes to handle overfitting more efficiently. Therefore, if the input variables (features) have a larger number compared to the number of training data, in some cases it can result in complex and ineffective models. Basically, the generalizability of the classifier may not be enough, being necessary to extract and select features to improve the generalizability. In recent decades, several feature extraction algorithms have been created and employed, being Principal Component Analysis (PCA) and Linear Discriminant Analysis (LDA) the most widely known \citep{Abdi2010}.

Parameter extraction methods assist in data pre-processing, and are later used to increase the performance of classical classification algorithms such as Support Vector Machine (SVM) \citep{Nagata2020}, Neural Network (NN) \citep{Xie2013}, Random Forest (RF) \citep{Zhang2003} and K-Nearest Neighbors (KNN) \citep{Pan}. Classification algorithms are widely used, however, they have some disadvantages, such as the dependence on auxiliary extraction algorithms to obtain satisfactory performance on certain types of data. Furthermore, the models resulting from the classifiers generally have low interpretability, i.e., understanding the relationship between the inputs and outputs of the predictor may not be simple. Even though classifiers are very efficient in solving several problems, the performance of a classifier depends on the nature of the data to be classified \citep{Silva2020a}. Since there is no single classifier that works best for all the problems provided.

Thus, this work proposes a hybrid algorithm that integrates machine learning and system identification methodologies in solving multinomial classification problems. One of the main motivations is the combination of methods and techniques from different areas in order to improve their performance. As a result, combinations are generated that usually provide more efficient hybrid algorithms. The approach allows to extract and select sets of representative features with reduced dimensionality, in the same way as it efficiently predicts categorical outputs. Furthermore, the algorithm allows the inclusion of lagged terms directly and deals with multicollinearity, resulting in more interpretable models, something that is not achievable using other popular classification techniques.

\section{Nonlinear System Identification}\label{sec_ind}

System Identification is an experimental approach that aims to identify and adjust a mathematical model of a system, based on experimental data that record the behavior of system inputs and outputs \citep{Billings2013,Aguirre}. In particular, the interest in nonlinear system identification has received a lot of attention from researchers since the 1950s and many relevant results have been developed \citep{Wei2004b,NM2016,Ferreira2017a}. A model representation constantly employed is the NARX model (\textit{Nonlinear AutoRegressive with eXogenous inputs}), consisting of a mathematical model based on differential equations. In general, system identification consists of several steps, including data collection and processing, choice of mathematical representation, determination of model structure, parameter estimation, and model validation \citep{Soderstrom1989}. For nonlinear systems, there are numerous techniques to determine the structure of the model, such as: Clustering Algorithms \citep{Jacome}, Genetic Programming \citep{Sette2001} and the method Orthogonal Forward Regression (OFR) using the Error Reduction Ratio (ERR) approach \citep{Wei2004b}.

\subsection{NARX Representation}

The NARX representation is a discrete-time model that explain the output value $y(k)$ as a function of previous values of the output and input signals:

\begin{equation}
\label{eqnarx}
\begin{aligned}
    y(k) = f^{l}(y(k-1),\cdots,y(k-n_y), \\
    u(k-1),\cdots,u(k-n_u))+e(k),
\end{aligned}
\end{equation}

\noindent where $f^l$ represents a nonlinear function of the model with nonlinearity degree $l \in \mathbb{N}$, $y(k)\in \mathbb{R}^{n_y}$ is the output of the system, and $u( k)\in \mathbb{R}^{n_u}$ is the input to the system in discrete time $k = 1, 2, \dots, N$; $N$ is the number of observations, $e(k)\in \mathbb{R}^{n_e}$ represents the uncertainties and possible noise in discrete time $k$, $n_y \in \mathbb{N}$ and $n_u \in \mathbb{N}$ describes the maximum lags for the output and input sequences, respectively. Most approaches assume that the function $f^l$ can be approximated by a linear combination of a predefined set of functions $\phi_i(\varphi(k))$, so that Equation (\ref{eqnarx}) can be expressed in the following parametric form:

\begin{equation}
\label{parnarx}
    y(k) = \sum_{i=1}^{m}\theta_{i}\phi_{i}(\varphi(k))+e(k),
\end{equation}

\noindent where $\theta_i$ are the coefficient to be estimated, $\phi_{i}(\varphi(k))$ are the predefined functions that depend on the regression vector:

\begin{equation}
\label{vetregnarx}
\begin{aligned}
    \varphi(k)=[y(k-1),\cdots,y(k-n_y), \\ u(k-1),\cdots,u(k-n_u)]^\mathsf{T},
\end{aligned}
\end{equation}

\noindent where $\varphi(k)$ represents the previous outputs and inputs, and $m$ is the number of functions in the set. One of the most used NARX models is the polynomial representation, where Equation (\ref{parnarx}) can be denoted as follows:

\begin{multline} 
\label{eqk}
y(k) = \theta_{0} + \sum_{i_1=1}^{n} \theta_{i_{1}} x_{i_{1}} (k) +  \sum_{i_1=1}^{n} \sum_{i_2=i_1}^{n} \theta_{i_{1}i_2}x_{i_1}(k)x_{i_2}(k)  + \\ \sum_{i_1=1}^{n} \cdots \sum_{i_l=i_l-1}^{n} \theta_{i_{1}i_{2} \cdots i_l}x_{i_1}(k) x_{i_2}(k) \cdots x_{i_l}(k) +e(k),
\end{multline}

\noindent considering $n=n_y+n_u$,

\begin{equation}
x_i(k) =  \begin{cases} y(k-i),  & \quad 1\le i \le n_y, \\ u(k-i   +n_y),  & \quad n_y+1 \le i \le n, \end{cases}     
\end{equation}

\noindent being $l$ the nonlinearity degree. The NARX model of order $l$ means that the order of each term in the model is not greater than $l$. The total number of potential terms in a polynomial NARX model is given by:

\begin{equation}
\label{eq_pot_termos}
    M = \frac{(n+l)!}{n! \cdot l!}.
\end{equation}

Finally, NARX models can be used to describe a wide variety of systems, simply obtaining analytical information about dynamic models. Another advantage is parsimony, meaning that a wide range of behaviors can be concisely represented using just a few terms from the vast search space formed by candidate regressors, as well as a small data set is needed to estimate a model, which can be crucial in applications where it is difficult to acquire a large amount of data.

\subsection{Orthogonal Forward Regression}

In general, the determination of the model structure and parameter estimation are performed together. One of the most popular algorithms for performing the two-step NARX modeling is the Orthogonal Forward Regression (OFR) algorithm \citep{Billings2013} . The algorithm transforms a set of candidate terms into orthogonal vectors and classifies them based on their contribution to the output data, identifying and fitting a deterministic and parsimonious NARX model that can be expressed in a form of generalized linear regression. The original Orthogonal Forward Regression algorithm uses the Error Reduction Rate (ERR) as a dependency metric. The criterion associates to each candidate term an index corresponding to the contribution in explaining the variance of the system output data. The error reduction rate is defined as the Pearson correlation coefficient $C(x,y)$ between two associated vectors $x$ and $y$:
\begin{equation}
\label{eq_err}
    C(x,y) = \frac{(x^{T}y)^2}{(x^{T}x)(y^{t}y)}.
\end{equation}

\section{Logistic-NARX Multinomial Classification }\label{sec_multi}  

Classification problems occur in the most diverse areas of knowledge, such as finance, healthcare, and engineering, where the goal is to identify a model that is capable of classifying observations or measurements into different categories or classes. A widely used approach is logistic regression, which uses concepts of statistics and probability to categorize variables by classes. In the logistic regression method, the predicted values are probabilities, so they are restricted to values between $0$ and $1$, and use the logistic function defined as:

\begin{equation}
    f(x) = \frac{1}{1+\exp{(-x)}},
\end{equation}

\noindent which $x \in \mathbb{R}$ and $f(x)$ is restricted to the range between $0$ and $1$. A problem found in logistic regression is multicollinearity, in which independent variables have exact or approximately exact linear relationships. If the variables are highly correlated, inferences based on the regression model may be erroneous or unreliable.

In this work, a hybrid multinomial classification method is presented, which allows the extraction and selection of features during the process. One of the advantages of using NARX modeling methodologies is the orthogonalization procedures, which address the multicollinearity problem by verifying the correlations between the predictor variables. The method is based on the Orthogonal Forward Regression \citep{AyalaSolares2017} algorithm that selects the terms and combines the logistic function with the NARX representation to obtain a probability model:

\begin{equation}
    p(x) = \frac{1}{1+\exp{\left[ \sum_{m=1}^{M}\theta_{m}\phi_{m} \left(\varphi(k) \right)  \right]}}.
\end{equation}

The multiclass problems are usually more complex than the binary classification, due to their decision boundaries. Direct extension of the binary algorithm to a multiclass version is not always possible or easy to accomplish. Therefore, the forms most explored by the scientific community are based on the binarization of multiclass problems. One of the most employed decomposition methods is One-Versus-All (OVA), which makes use of $C$ binary classifiers to solve a classification problem involving $C$ classes. For the $v$-th binary classifier, a distinction is made between the class $w_v$ and the other classes. In this way, a $x$ pattern is classified by the following decision rule:
\begin{equation}
\label{eq_multi}
    x \in w_v \Leftrightarrow \underset{1 \leq v \leq C }{\arg\max} \ f_v (x),
\end{equation}

\noindent defining $f_v$ as the result given by the model referring to the class $v$, meaning the probability between $0$ and $1$ of belonging to the v-th class with respect to an instance $x$. Then, it is verified which class is most likely given as a result in Equation (\ref{eq_multi}).

\begin{englishalgorithm}
\caption{Logistic-NARX Multinomial\label{alg_1}}
\begin{algorithmic}[1]
    \State \textbf{Input:} $\{y(k), k = 1, \dots, N \}$, $\mathcal{M} = \{\phi_i, i = 1, \dots, m\}$ , $l$, $n_y$, $n_u$, $k$ 
    \State \textbf{Output:} $\boldsymbol{\alpha} = \{\alpha_i, i = 1, \dots, k\}$, $\boldsymbol{\theta} = \{ \theta_i, i = 1,\dots,k\}$ 
    \For{$i=1:m$}
         \State $w_i \gets \frac{\phi_{i}}{\left \| \phi_{i} \right \|_{2}}$ 
        \State $r_{i} \gets $ Logistic regression accuracy in $w_i$ and $y$
         \EndFor
    \State $ j \gets \text{arg} \underset{1 \leq i \leq m }{\max}\{ r(w_{i},y) \}$ 
    \State $q_1 \gets w_j$
    \State $ \alpha_1 \gets \phi_{j}$
    \State Train logistic model with $\alpha_1$ and $y$
    \State Compute cross-validation
    \State Remove $\phi_j$ from $\mathcal{M}$
    \For{$s=2:k$}
         \For{$i=1:m$}
             \State $w_{i}^{(s)} \gets$ Orthogonalize $\phi_i$ in $[q_{_1},\dots,q_{_{(s-1)}}]$
             \If{$w_i^{\mathsf{T}}w_i<10^{-10}$}
             \State Remove $\phi_i$ from $\mathcal{M}$\;
             \State Next iteration\;
             \EndIf
             \State $r_{i} \gets $ Logistic regression accuracy in $w_i$ and $y$
         \EndFor 
         \State $ j \gets \underset{1 \leq i \leq m-s+1 }{\max}\{ r^{(i)}(w_{i},y) \}$ 
         \State  $q_s \gets w_j$
         \State  $ \alpha_s \gets \phi_{j}$
         \State Remove $\phi_j$ from $\mathcal{M}$
         \State $\boldsymbol{\alpha} \gets [\alpha_1,\dots,\alpha_{_{(s)}}]$
         \State Train logistic model with $\boldsymbol{\alpha}$ and $y$
         \State Compute cross-validation
    \EndFor 
    \State $\boldsymbol{\alpha} \gets [\alpha_1,\dots,\alpha_{_{(k)}}]$ \Comment{matrix of selected terms}
    \State $\boldsymbol{\theta} \gets [\theta_1,\dots,\theta_{_{(k)}}]$ \Comment{estimated coefficients vector}
\end{algorithmic}
\end{englishalgorithm}

\begin{table}[b!]
\caption{Summary of the datasets.}\label{tb_model_1}
\begin{center}
\centering
\setlength{\tabcolsep}{15.5pt} 
\renewcommand{\arraystretch}{1.5}
\begin{tabular}{lccc}
\hline
Dataset &  Classes & Features & Samples  \\\hline
Iris &  3 &  4 &  150  \\ \hline
Wine & 3 &  13 & 178  \\ \hline
Glass & 6 &  9 & 214 \\ \hline
Wave & 3 &  40 & 5000  \\ \hline
\end{tabular}
\end{center}
\end{table}

To combine the methodologies of NARX models and multinomial classification, some aspects of the Orthogonal Forward Regression algorithm were adapted. The OFR algorithm relies on the error reduction rate given by Equation (\ref{eq_err}) to determine the contribution of each candidate term. However, this metric is no longer useful as the output is a categorical variable due to the multiclass addressed. To solve this problem, a simple logistic model using maximum likelihood estimate (MLE) is used. Basically, the accuracy of the prediction of the categorical variable resulting from the logistic model based on continuous variables is calculated. Since the resulting predictor has a high degree of accuracy, it can be concluded that the two variables are correlated. To calculate the accuracy, the K-Fold Cross Validation was used, in order to assess the generalizability of the model.

\begin{figure}[t!]
\hspace*{-0.45cm}  
\centering
\includegraphics[scale=0.235]{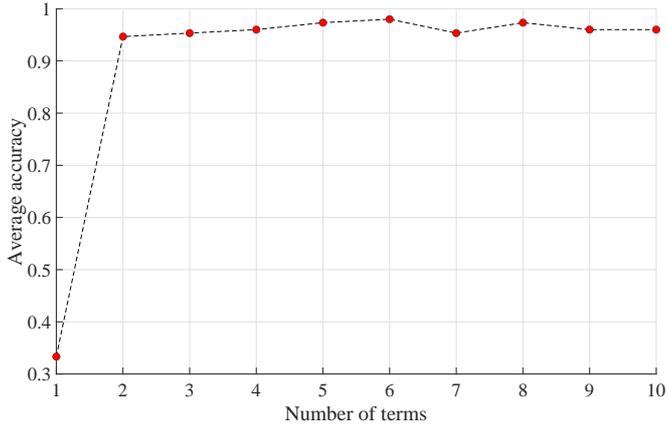}    
\caption{Relation between average accuracy and the number of selected terms associated with \textit{Iris} data.} 
\label{fig_ac_1}
\end{figure}

 In Algorithm \ref{alg_1}, line ($1$) represents the inputs composed by the vector $y(k)$ of classes (labels), the matrix $\mathcal{M}$ constitutes the regressors formed by combinations of feature vectors, $k$ is the maximum number of selected terms and ($l$, $n_y$, $n_u$) are the parameters of the NARX model (Equation \ref{eqnarx}). Lines (3-8) aim to select the candidate terms $\phi_i$ with greater discriminatory power, based on the prediction accuracy of the logistic model. In lines (9-10) the logistic model is trained using the selected regressors $\alpha$ and the class vector $y(k)$, calculating the cross-validation of the classification. The selected terms are transformed at each step into a new group of orthogonal bases in lines (14-18), using the Gram-Schmidt Orthogonalization procedures. The process is repeated on lines (12-25) until it reaches the specified maximum number of model terms selection $k$. Finally, in lines (26-27) the matrix of selected terms $\boldsymbol{\alpha}$ and the vector of coefficients $\boldsymbol{\theta}$ of the estimated model are obtained. Since the number of terms in the model is not known in advance, the parameter $k$ can be selected heuristically, executing the Algorithm \ref{alg_1} and checking the resulting accuracy curve.
 
 \vspace{0.5cm}
\begin{figure}[t!]
\hspace*{-0.45cm}  
\centering
\includegraphics[scale=0.235]{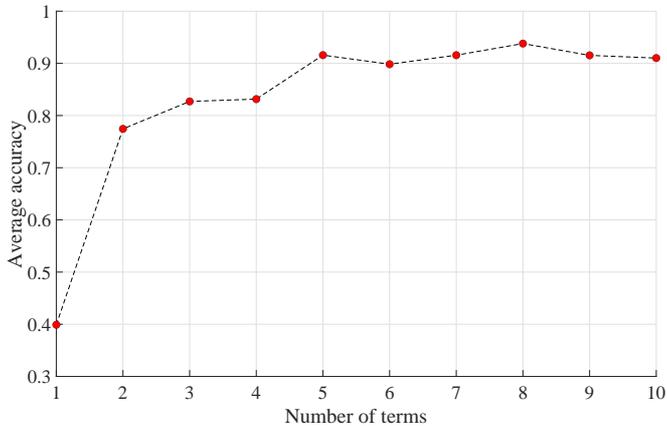}    
\caption{Relation between average accuracy and the number of selected terms associated with \textit{Wine} data.} 
\label{fig_ac_2}
\end{figure}

\section{Results}

In this section, simulations are carried out to evaluate the dimensionality reduction capacity and classification accuracy of the proposed method, with tests being carried out in classical databases in the literature. Furthermore, the effectiveness of the new methodology in comparison to popular classification methods will be analyzed.

\begin{figure}[t!]
\hspace*{-0.45cm}  
\centering
\includegraphics[scale=0.235]{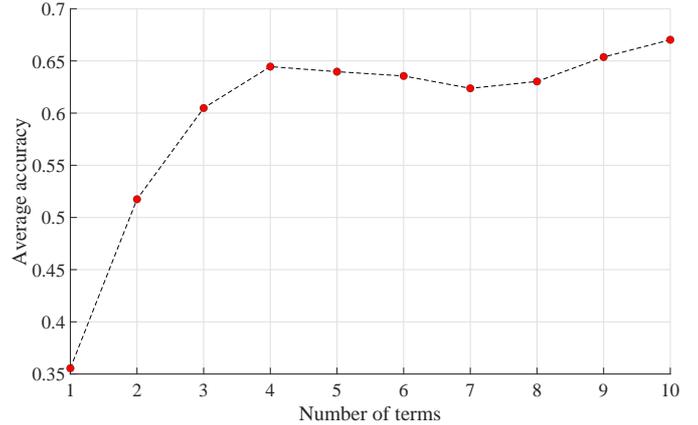}  
\caption{Relation between average accuracy and the number of selected terms associated with \textit{Glass} data.} 
\label{fig_ac_3}
\end{figure}

\begin{figure}[t!]
\hspace*{-0.45cm}  
\centering
\includegraphics[scale=0.235]{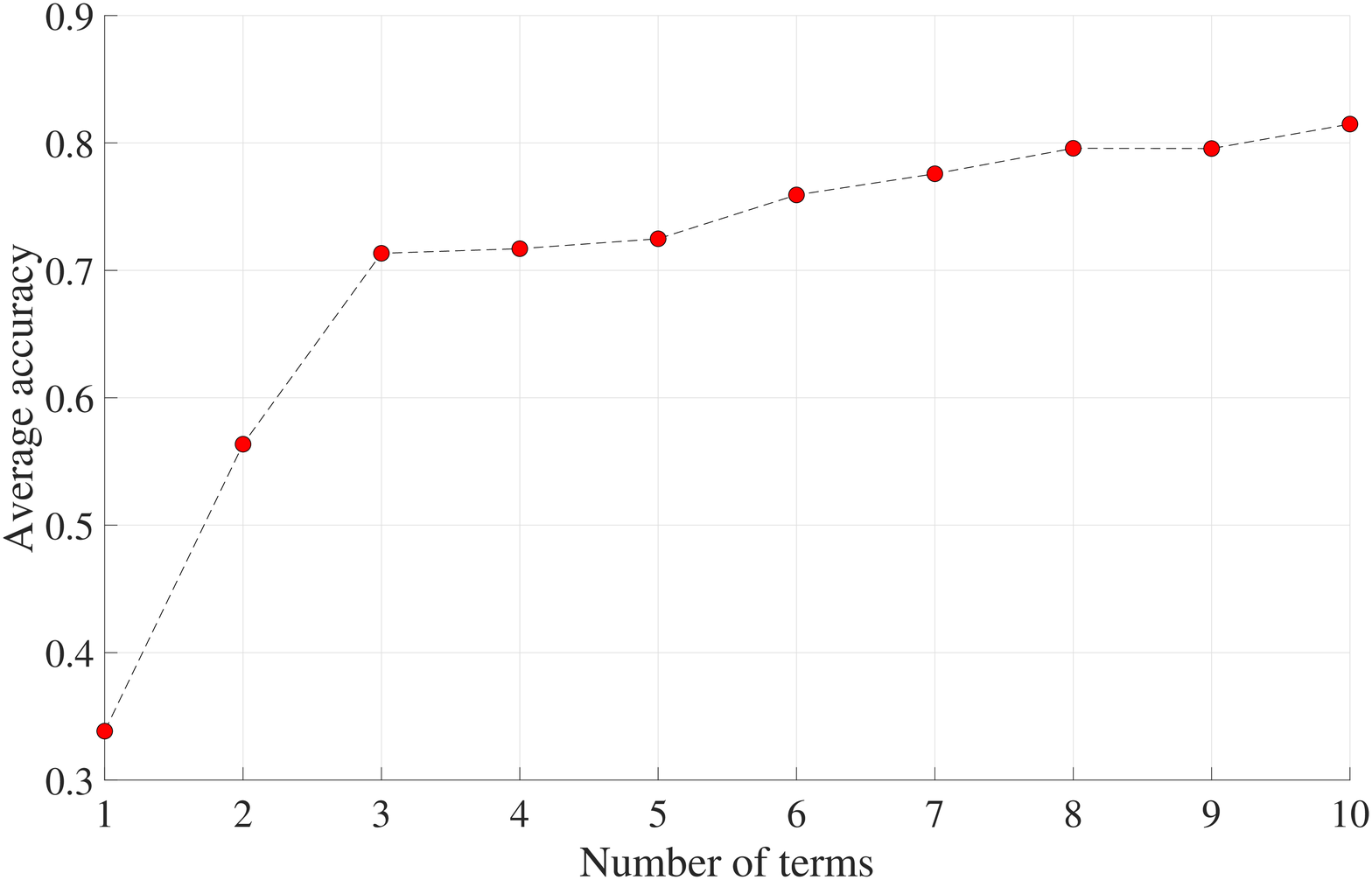}   
\caption{Relation between average accuracy and the number of selected terms associated with \textit{Wave} data.} 
\label{fig_ac_4}
\end{figure}

In order to analyze the performance of the algorithm, four multivariate data sets available in the UCI Machine Learning\footnote{Repository of Iris, Wine, Glass and Wave datasets (UCI Machine Learning Repository): \url{https://archive.ics.uci.edu/ml}.}were selected, which are often studied in machine learning. The choice of data sets was based on the range of diversity, considering the number of features, samples, classes, and nature of the data. The \textit{Iris} dataset is a classic example from the literature, which has simple discrimination. \textit{Wine} data has more features for analysis and has imbalanced classes. The \textit{Glass} problem is composed of $6$ classes and their sets have minority classes and data constituted as outliers. Finally, \textit{Wave} sets have a large number of features and samples composed of noise. A summary of the datasets is shown in Table \ref{tb_model_1}.

\newcolumntype{g}{>{\columncolor{lightgray}}c}
\begin{table*}[t]
\caption{Selected model terms and score values for each dataset.}\label{tb_com_3}
\begin{center}
\setlength{\tabcolsep}{7.6pt} 
\renewcommand{\arraystretch}{1.5}
\begin{tabular}{ggccggcc}
\hline
 \multicolumn{2}{g}{Iris}  & \multicolumn{2}{c}{Wine}   & \multicolumn{2}{g}{Glass} & \multicolumn{2}{c}{Wave} \\ \hline 
 Terms & Score & Terms & Score & Terms & Score & Terms & Score \\ \hline 
 $u_{3}(k-1)$ & $0.9533$  & $\text{Constant}$ & $0.7759$  & $\text{Constant}$ & $0.7287$ & $\text{Constant}$ & $0.7690$ \\ \hline
$\text{Constant}$ & $0.7714$ & $u_{7}(k-1)$ & $0.7695$  & $u_{4}(k-1)$ & $0.5183$ &  $u_{7}(k-1)$ & $0.5632$ \\ \hline
$u_{4}(k-1)u_{4}(k-1)$  & $0.6067$ & $u_{10}(k-1)u_{13}(k-1)$ & $0.6680$ & $u_{3}(k-1)u_{7}(k-1)$ & $0.5434$ & $u_{11}(k-1)$ & $0.5386$ \\ \hline 
$u_{3}(k-1)u_{3}(k-1)$  & $0.4933$ & $u_{13}(k-1)$ & $0.6527$  & $u_{3}(k-1)u_{3}(k-1)$  & $0.4814$ & $u_{8}(k-1)u_{11}(k-1)$ & $0.4518$ \\ \hline
$u_{2}(k-1)u_{4}(k-1)$ & $0.4733$ & $u_{7}(k-1)u_{11}(k-1)$ & $0.5189$ & $u_{4}(k-1)u_{6}(k-1)$ & $0.4773$ & $u_{12}(k-1)u_{16}(k-1)$ & $0.4264$\\ \hline
$u_{1}(k-1)u_{4}(k-1)$  & $0.4067$ & - & - & $u_{3}(k-1)u_{4}(k-1)$ & $0.4399$ &  $u_{10}(k-1)$ & $0.4266$ \\ \hline
- & - & - & - & - & - & $u_{15}(k-1)u_{16}(k-1)$ & $0.4286$ \\ \hline
- & - & - & - & - & - & $u_{16}(k-1)$ & $0.4142$ \\ \hline
- & - & - & - & - & - & $u_{5}(k-1)u_{10}(k-1)$ & $0.4052$ \\ \hline
- & - & - & - & - & - & $u_{16}(k-1)$ & $0.4014$  \\ \hline
\end{tabular}
\end{center}
\end{table*}

The algorithms were implemented in Matlab and executed using a machine with Intel Core i5, CPU $2.5 \ \text{GHz}$ and $8 \ \text{Gb} \ \text{RAM}$. In all methods presented, the same datasets of training ($80\%$) and validation ($20\%$) were applied, employing the $5$-fold cross-validation. The features of the data sets were normalized to zero mean and standard deviation equal to 1, being the only preprocessing performed. In the classification using the proposed method, the following parameters were considered: degree of nonlinearity $l=2$, maximum lags $n_u=n_y=2$ and maximum number of selected terms $k=10$. For comparison purposes, the parameters inserted in the classification methods were chosen based on tests with the databases, selecting the configurations with greater accuracy using $5$-fold cross-validation. The configuration used in each dataset is presented below:

\begin{itemize}
    \item Iris - (RF) with division criterion Gini's diversity index and maximum number of decision divisions equal to $5$, (SVM) with Polynomial kernel and order $2$ and (KNN) with metric Minkowski distance using exponent equal to $3$ and number of nearest neighbors equal to $10$;
    \item Wine - (RF) with division criterion Gini's diversity index and maximum decision divisions equal to $20$, (SVM) with Polynomial kernel and order $2$ and (KNN) with metric Euclidean distance and number of nearest neighbors equal to $10$;
    \item Glass - (RF) with division criterion Gini's diversity index and maximum decision divisions equal to $100$, (SVM) with Polynomial kernel and order $3$ and (KNN) with metric Euclidean distance and number of nearest neighbors equal to $10$;
    \item Wave - (RF) with division criterion Gini's diversity index and maximum decision divisions equal to $20$, (SVM) with Gaussian kernel and (KNN) with metric Euclidean distance and nearest neighbors equal to $10$.

\end{itemize}

\begin{table}[b!]
\caption{Comparison between average and maximum accuracy resulting from cross-validation.}\label{tb_model_2}
\begin{center}
\centering
\setlength{\tabcolsep}{9.1pt} 
\renewcommand{\arraystretch}{1.5}
\begin{tabular}{lccccc}
\hline
&  & Iris & Wine & Glass & Wave \\\hline
\multirow{2}{*}{NARX} &  $\overline{x}$  &  \cellcolor{gray!10}0.9800 & \cellcolor{gray!10} 0.9382 & \cellcolor{gray!10}0.6682 & \cellcolor{gray!10}0.8148\\ 
 & max &  \cellcolor{gray!10}1.0000 & \cellcolor{gray!10}1.0000 & \cellcolor{gray!10}0.7907 & \cellcolor{gray!10}0.8228 \\ \hline
\multirow{2}{*}{RF} & $\overline{x}$  &  0.9467 & 0.8427 & 0.6449 & 0.7448 \\ 
 & max &  0.9680 & 0.9189 & 0.7209 & 0.7538 \\ \hline
\multirow{2}{*}{SVM} & $\overline{x}$  & \cellcolor{gray!10} 0.9533 & \cellcolor{gray!10}0.9213 & 0.6682 & \cellcolor{gray!10}0.8650 \\ 
 & max & \cellcolor{gray!10} 1.0000 & \cellcolor{gray!10}0.9730 & 0.7727 & \cellcolor{gray!10}0.8749 \\ \hline
\multirow{2}{*}{KNN}& $\overline{x}$  & \cellcolor{gray!10} 0.9533 & 0.8652 & \cellcolor{gray!10}0.7150 & 0.7944 \\ 
 & max & \cellcolor{gray!10} 1.0000 & 0.9412 & \cellcolor{gray!10}0.7674 & 0.8208 \\ \hline
\end{tabular}
\end{center}
\end{table}

Figure \ref{fig_ac_1} represents the application of the algorithm to the Iris dataset. The results suggest that the selection of $2$ terms is sufficient to represent the classifier model, obtaining an average accuracy of $0.94$. Furthermore, selecting more terms did not significantly increase the accuracy. On the other hand, in Figure \ref{fig_ac_2} that represents the results applied to the Wine data, the selection of terms gradually increased the accuracy. Therefore, $5$ terms were chosen to represent the classifier model, resulting in an average accuracy of $0.91$. In Figure \ref{fig_ac_3} it is possible to observe $2$ local maximums in $4$ and $10$ terms, showing that in some cases the increase in the number of terms can result in an accuracy decrease. The simulation associated with the Wave data resulted in increased accuracy in the addition of new terms (see Figure \ref{fig_ac_4}), indicating that the maximum selection of $k=10$ can be further increased for greater accuracy. In summary, the method was successful in selecting the features, ranking the most relevant in the classification process.

\begin{table}[h]
\caption{Performance of the proposed method in feature extraction.}\label{tb_com_2}
\begin{center}
\setlength{\tabcolsep}{7.6pt} 
\renewcommand{\arraystretch}{1.5}
\begin{tabular}{lccccc}
\hline
\multirow{2}{*}{\begin{tabular}[c]{@{}l@{}}Dataset\end{tabular}} & \multicolumn{2}{c}{Features}   & \multirow{2}{*}{\begin{tabular}[c]{@{}c@{}}Reduction \\ (\%) \end{tabular}}  & \multicolumn{2}{c}{Accuracy}\\ \cline{2-3} \cline{5-6}
 & Data & Model &  & $\overline{x}$ & max \\ \hline
Iris & 4 & 2 & 50.00 & 0.9467 & 0.9667  \\ \hline 
Wine & 13 & 5 & 61.53 & 0.9158 & 0.9444 \\ \hline
Glass & 9 & 4 & 55.55 & 0.6445 & 0.7381 \\ \hline 
Wave & 40 & 10 & 75.00 & 0.8148 & 0.8228\\ \hline
\end{tabular}
\end{center}
\end{table}

Table $2$ represents the selected terms and their importance score for the classifier model using the proposed method. The values found explain the curves obtained in relation between accuracy ratio and the number of terms, since the growth rate is higher in the insertion of the most significant terms and degrades in the inclusion of the least significant terms (see Figure \ref{fig_ac_3}). The comparison between the average and maximum accuracy between the classification methods is shown in Table \ref{tb_model_2}. The method obtained higher average accuracy in the tests of the sets of Iris and Wine. While the KNN and SVM methods achieved better results in the Glass and Wave sets respectively. However, the proposed method obtained better results compared to the RF and SVM methods in the Glass set, as well as surpassing the RF and KNN methods in the Wave sets. Another important aspect is that the method in the Wave sets used only $10$ terms or features compared to the others that used $40$. The purpose of the comparison is to reveal that the method competes with other classical algorithms, making it an alternative that can obtain gains in certain data sets.

The performance of the proposed method in feature extraction, together with the dimensionality reduction capability, is summarized in Table \ref{tb_com_2}. The algorithm proposed in the tests significantly reduced the size of the data, resulting in average accuracy similar to those obtained by the other methods (see Table \ref{tb_model_2}). In emphasis, the \textit{Wine} test reduced the dimension by $61.53\%$, maintaining an average accuracy of $0.9158$ and resulting in greater accuracy than those found in RF and KNN in Table \ref{tb_model_2}. Likewise, in the \textit{Wave} data there was a reduction of $75 \%$ and an average accuracy of $0.8148$, consisting of better results compared to the RF and KNN techniques.

\section{Conclusion}

In this work, a hybrid technique was proposed that incorporates system identification and machine learning methodologies in the prediction of categorical variables. The presented algorithm performs the extraction and selection of features, ordering the most significant terms to compose a multiclass classification model. The model obtained is relatively simple and intuitive to interpret, providing insights with reduced dimensionality that clearly explain the incremental impact of a predictor variable on the response variable. The results show that the method excels in maximum accuracy by $3$ of the applied case studies. In terms of average accuracy, the method obtained better results in $2$ case studies. However, the proposed technique reduced the dimensionality of the data in the analyzed sets by more than $50 \%$, keeping the accuracy equivalent to the other approached techniques. In general, the method is an interesting alternative for extraction and classification, and can achieve significant gains in certain datasets. The results are promising and in future proposals we want to evaluate the method in comparison with other extraction and feature selection techniques. Another approach is to use heuristic methods in the term selection process, to obtain optimal values for the number of terms in the model concerning the average accuracy.


    
    
    
    
    
    
    
    
    

\section*{Acknowledgements}
We thank CAPES, CNPq, INERGE, FAPEMIG and
Federal University of Juiz de Fora (UFJF) for the support.

\bibliography{references}             

\end{document}